\documentclass[11pt,a4paper]{article}
\usepackage[hyperref]{acl2020}
\usepackage{times}
\usepackage{latexsym}
\usepackage{graphicx}
\usepackage{float}
\usepackage{multirow}
\usepackage{amsmath}
\usepackage{mathrsfs}

\usepackage{microtype}

\aclfinalcopy 

\title{AR: Auto-Repair the Synthetic Data for Neural Machine Translation}

\author{Shanbo Cheng, Shaohui Kuang, Rongxiang Weng \\ Heng Yu, Changfeng Zhu, Weihua Luo \\
\{shanbo.csb, shaohui.ksh, rongxiang.wrx\}@alibaba-inc.com}

\date{}

\begin{document}
\maketitle
\begin{abstract}


Compared with only using limited authentic parallel data as training corpus, many studies have proved that incorporating synthetic parallel data, which generated by back translation (BT) or forward translation (FT, or self-training), into the NMT training process can significantly improve translation quality. 
However, as a well-known shortcoming, synthetic parallel data is noisy because they are generated by an imperfect NMT system. As a result, the improvements in translation quality bring by the synthetic parallel data are greatly diminished. In this paper, we propose a novel Auto-Repair (AR) framework to improve the quality of synthetic data. Our proposed AR model can learn the transformation from low quality (noisy) input sentence to high quality sentence based on large scale monolingual data with BT and FT techniques. The noise in synthetic parallel data will be sufficiently eliminated by the proposed AR model and then the repaired synthetic parallel data can help the NMT models to achieve larger improvements. 
Experimental results show that our approach can effective improve the quality of synthetic parallel data and the NMT model with the repaired synthetic data achieves consistent improvements on both WMT14 EN$\rightarrow$DE and IWSLT14 DE$\rightarrow$EN translation tasks.
\end{abstract}

\section{Introduction}
\label{sec:introduction}
Neural machine translation (NMT) based on the \textit{encoder-decoder framework with attention mechanism}~\cite{sutskever2014sequence,bahdanau2014neural,Cho2014Learning,vaswani2017attention} has achieved state-of-the-art (SOTA) results in many language pairs~\cite{xia2019microsoft}. Generally, millions of or even more parallel sentence pairs are needed to train a decent NMT system~\cite{wu2016google}. However, authentic parallel data-set is limited in many scenarios, e.g. low resource language, which restricts the use of NMT in the real world~\cite{sennrich2019revisiting}.

As collecting large scale authentic parallel data is expensive and impractical in many scenarios, approaches that use freely available monolingual data to create an additional synthetic parallel data have drawn much attention, e,g back translation. Since the pioneer work of~\cite{sennrich2017university} uses training data consists of synthetic and authentic parallel data to train a high quality NMT model, many  approaches~\cite{gulcehre2015using,he2016dual,poncelas2018investigating,lample2018phrase,edunov2018understanding} have been proposed to further proved that synthetic parallel data obtained by back-translation (BT) or forward-translation (FT)~\cite{bogoychev2019domain} is a simple yet effective approach to improve translation quality.

Most of the existing work focuses on how to better leverage given synthetic data. 
\newcite{caswell2019tagged} proposed tagged back-translation to guide the training process by informing the NMT model that the given source is
synthetic. 
The work in~\newcite{wang2019improving} uses pre-calculated uncertainty scores for BT data to weight the attention of NMT models. 
\newcite{wu2019exploiting} proposed to use noised training to better leverage both BT and FT data. 
Despite the success of above works, they haven't propose any method to improve the quality of synthetic data while the quality of synthetic data matters in NMT training~\cite{gulcehre2015using}.
Iterative BT~\cite{hoang2018iterative,gwinnup2017afrl} generates increasingly better synthetic data along with the training iterations. However, the iteration procedure is time-consuming and expensive in real-world applications.

\begin{figure}[h]
	\centering
	\includegraphics[height=1.1in,width=3in]{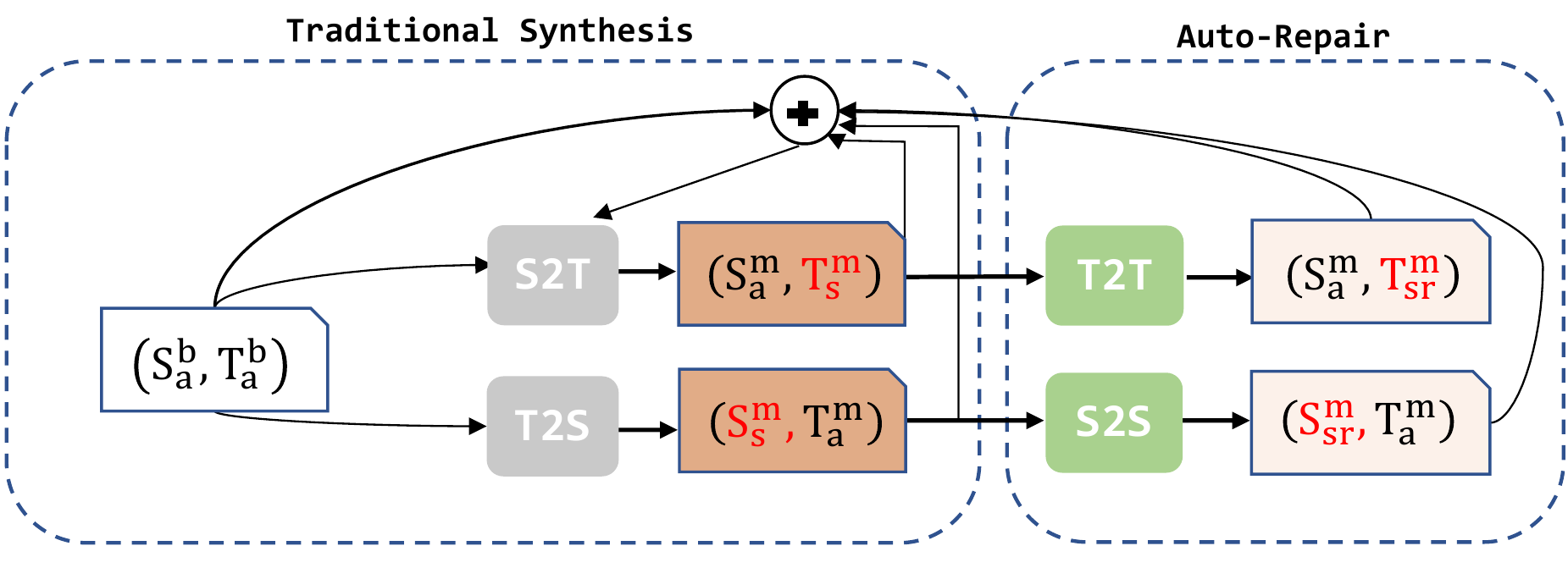} 
	\caption{An overview of our AR training framework.}
	\label{fig:strategy}
\end{figure}

In this paper, we propose a simple and effective auto-repair (AR) model to improve the quality of synthetic parallel data with less time cost and can be applied to various types of synthetic corpus. 
Our auto-repair model could learn the mapping from low quality synthetic data to high quality authentic data. 
After the training, the auto-repair (AR) model can be used to repair the mistakes in synthetic data directly.
NMT models trained by the repaired synthetic data can achieve more gain compared with original back-translated synthetic data.
%
We evaluate the proposed approach on WMT14 EN$\rightarrow$DE and IWSLT14 DE$\rightarrow$EN translation tasks.
Experimental results show that our AR model could repair the mistakes in synthetic data effectively. And the translation quality of both tasks get considerable improvements with the high quality synthetic data.

\section{Approach}
\label{sec:approach}
In this section, we elaborate on the proposed AR framework and how we integrate AR models into NMT. 

\subsection{Background and Notation}
\label{sec:background}
We firstly define $S_a$ and $T_a$ as source authentic sentence and target authentic sentences, respectively. Meanwhile, $S_s$ and $T_s$ as source and target synthetic sentence, respectively. Since our data come from either bilingual or monolingual data, we use different superscripts to indicate the data source. For example, $S_a^b$ and $S_a^m$ represent the source authentic sentence from bilingual data and monolingual data, respectively. 

The left part (Traditional Synthesis) of Figure~\ref{fig:strategy} shows a diagram of a traditional data synthesis process. We firstly pre-train a source-to-target \textbf{S2T} and a target-to-source \textbf{T2S} NMT model based on the authentic parallel corpus $(S_a^{b}, T_a^{b})$. Then we use the pre-trained models to translate the source and target side \textit{monolingual} data, $S_a^{m}$ and $T_a^{m}$, to get the target synthetic sentence $T_s^{m}$ and source synthetic sentence $S_s^{m}$, respectively. Finally, we can use the combination of authentic data $(S_a^{b}, T_a^{b})$, FT pseudo-parallel corpus $(S_a^{m}, T_s^{m})$ and BT pseudo-parallel $(S_s^{m}, T_a^{m})$ to train our S2T model.


%

\subsection{The Proposed AR Framework}
\label{sec:pcs}
The right part (Auto-Repair) of Figure~\ref{fig:strategy} schematically illustrates the AR Framework. We follow the definitions for traditional synthesis in Section~\ref{sec:background}, and we define $S_{sr}^{m}$ and $T_{sr}^{m}$ as the repaired synthetic source sentence and target sentences, respectively. Rather than directly use the synthetic data, ($S_a^{m}, T_s^{m}$) and ($S_s^{m}, T_a^{m}$), we use a $T2T$ and a $S2S$ auto-repair model to \textit{repair} the synthetic data and get the ($S_a^{m}, T_{sr}^{m}$) and ($S_{sr}^{m}, T_a^{m}$) data. Then we continue to train the \textbf{S2T} model with the combination of authentic parallel corpus $(S_a^{b}, T_a^{b})$, FT synthetic (pseudo-parallel) corpus $(S_a^{m}, T_s^{m})$, BT synthetic $(S_s^{m}, T_a^{m})$, FT repaired corpus $(S_a^{m}, T_{sr}^{m})$ and BT repaired corpus $(S_{sr}^{m}, T_a^{m})$.
For simplicity reasons, we drop the superscripts of symbolic representation in the following descriptions for monolingual data. e.g $S_a$ for $S_a^m$.
\paragraph{Auto-Repair Modeling}
The aim of the AR model is to transform the low quality (noisy) sentences to high quality sentences.
We adopt the seq2seq architecture to build our \textit{S2S} and \textit{T2T} AR models. In this work, we use the SAN-based (self-attention network)~\cite{vaswani2017attention} structure. 
Given a low and high quality sentence-pair $(s_s, s_{sr})$, where $s_s = (s_1, s_2, ..., s_m)$ and a high quality sentence $s_{sr} = (sr_1, sr_2, ..., sr_n)$,  the conditional distribution of each target token $p(sr_i)$ predicted by AR model is computed as 
\begin{equation}
p(sr_i|sr_{<i},s_s) = AR(sr_i|sr_{<i},s_s))
\end{equation}
The input of our AR model is $s_s$, and the AR model transforms it to $s_{sr}$, which is of higher quality than $s_s$. 

\begin{table*}[!t]
\centering
\begin{tabular}{l||cc|cc|cc}
\hline
\multirow{2}{*}{Model}             & \multicolumn{2}{c|}{WMT14 EN$\rightarrow$DE}   & \multicolumn{2}{c}{IWSLT14 DE$\rightarrow$EN} & \multicolumn{2}{|c}{AVG}  \\ &  BLEU &$\Delta$    &   BLEU &$\Delta$ &    BLEU &$\Delta$     \\ \hline
Transformer-base~\cite{vaswani2017attention} & 27.3&N/A&    N/A             & N/A  & N/A & N/A \\
Transformer-base~\cite{gao2019soft}& N/A & N/A&34.79& N/A & N/A & N/A \\\hline
Transformer-base (BASE) & 27.52 & N/A & 35.04 & N/A & 31.28 & N/A \\\hline
BASE + BT~\cite{sennrich2017university}     & 28.64  & +1.12 & 37.11 & +2.07 & 32.88 & +1.60 \\
BASE + FT                         & 27.88   &  +0.36  & 35.88 & +0.84 &    31.88 & +0.60\\
BASE + BT + FT                      & 28.84  &  +1.32  & 37.70 & +2.66 &   33.27 & +1.99 \\
\hline
BASE + BTR-REP                 & 29.05  &  +1.53 & 37.79 & +2.75 & 33.42 & +2.14  \\
BASE + FTR-REP                 & 28.11 &   +0.59  & 36.17 & + 1.13 & 32.14 & +0.86 \\
BASE + BTR-ADD                 & 29.29 &  +1.77  & 38.15 & +3.11 & 33.72 & +2.44 \\
BASE + FTR-ADD                & 28.20   & +0.68  & 36.34 & +1.30 & 32.27 & +0.99  \\
BASE + BTR-ADD + FTR-ADD  & \textbf{29.59}  &  \textbf{+2.07} & \textbf{38.89} & \textbf{+3.85} & \textbf{34.24} & \textbf{+2.96}  \\ 
\hline
\end{tabular}
\caption{The translation quality results on WMT14 and IWSLT14 tasks. \textbf{+BT} and \textbf{+FT} means adding BT data and FT data to the bilingual base corpus, respectively. 
\textbf{BTR} and \textbf{FTR} means the repaired BT data and repaired FT data, respectively.
\textbf{BTR-REP} means replacing BT data with BTR data, and adding the BTR data to the bilingual base corpus.
\textbf{BTR-ADD} means adding both BT and BTR data to the base corpus. \textbf{FTR-REP} and \textbf{FTR-ADD} shares the similar definition with \textbf{BTR}. The ratio of authentic data to synthetic data is 1:1.}
\label{table:trans}
\end{table*}

\paragraph{AR Model Training}
\label{sec:artd}
We take BT scenario as an example to describe how we generate the training data and how we train our AR models. The FT scenario is identical to BT, except the language to be repaired is different. In order to build the AR model for BT data, we need sentence-pair $(S_s, S_{sr})$ as training corpus, where $S_s$ is generated by NMT systems and of low quality. $S_{sr}$ is high quality sentence.

For the $S_{sr}$, we could simply use the large scale authentic monolingual data $S_a$ for two reasons: 1) the monolingual data is almost always universally available~\cite{gulcehre2015using}, which makes it abundant for NMT training; 2) the monolingual data is original in the specific language, thus the fluency and accuracy are guaranteed, which makes it of high quality. \newline
For the $S_s$, we investigate a data-driven method to generate it. We firstly use the pre-trained \textbf{S2T} model to translate the monolingual data, $S_a$, and get $T_s$. Then we use the pre-trained \textbf{T2S} model to translate from $T_s$ to $S_s$.



Because the $S_s$ here is generated by \textbf{S2T} and \textbf{T2S} NMT models, it could reveal the mistakes made by NMT models inherently, which exactly meets our requirements for $S_s$. In addition, the AR dev data contains 1000 sentence pairs randomly selected from AR training data. After the training and dev data are obtained, we can train the AR models following the typical seq2seq model training methods~\cite{vaswani2017attention}.

\section{Experiments}
\label{sec:experiments}

\subsection{Implementation Detail}
\paragraph{Data-set}
We evaluate the proposed approach on the publicly available WMT2014 English to German (EN$\rightarrow$DE) and IWSLT2014 German to English (DE$\rightarrow$EN) tasks. On the WMT EN$\rightarrow$DE task, our training set consists of about 4.5 million sentence pairs. We use \texttt{newstest2013} as our validation set and \texttt{newstest2014} as our test set\footnote{http://www.statmt.org/wmt17/translation-task.html}. 
We learned a joint byte pair encoding (BPE) ~\cite{sennrich2015neural} for English and German with 32,000 operations. We limit the size of our English and German vocabularies to 30,000. 
On the IWSLT14 DE$\rightarrow$EN task, following the setting of \newcite{gehring2017convolutional}, the training set includes about 160k sentence pairs and the dev set includes 7000 sentence pairs randomly selected from the training set. Then, we concatenate \texttt{tst2010$\sim$12}, \texttt{dev2010} and \texttt{dev2012} as our test data\footnote{https://github.com/facebookresearch/fairseq/tree/master/data}.
We learned a joint BPE with 10,000 operations and limit the size of both English and German vocabularies to 10,000. 
We randomly sampled 16 million monolingual sentences for both English and German from WMT News Crwal data for both tasks. 

\paragraph{Settings}
We use an in-house implementation of Transformer~\footnote{https://github.com/tensorflow/tensor2tensor/tree/master/ tensor2tensor}. 
For the Transformer model structure, we use the transformer-base settings from~\cite{vaswani2017attention}. 
Sentence pairs are batched together by approximate sentence length. Each batch has approximately 25000 source tokens and 25000 target tokens.
We set our label smoothing and dropout rate to 0.1. We use the Adam~\cite{kingma2014adam} to update the parameters, and the learning rate was varied under a warm-up strategy with 4000 steps.
We use beam search for heuristic decoding, and the beam size is set to 4 and case-insensitive 4-gram BLEU~\cite{papineni2002bleu} to evaluate our translation results.



\subsection{Experimental Results}
\label{ssec:sda}

We show our translation results on WMT14 EN$\rightarrow$DE task and IWSLT14 DE$\rightarrow$EN task in Table~\ref{table:trans}. The first 3 lines show that our in-house implementation is comparable with the open source implementation~\cite{vaswani2018tensor2tensor}.

\paragraph{Results of FT \& BT} The next 3 lines of Table~\ref{table:trans} show the consistent improvements when using different synthetic data. The BASE + BT + FT model outperforms the baseline by an average 1.99 BLEU points and outperforms the other two models (+FT or +BT) by 0.4$\sim$1.39 BLEU points on the two tasks. It's worth mentioning that the translation quality improvement bring by FT \footnote{We found that directly adding FT data into authentic bilingual data and continue to train the NMT model, the translation quality deteriorates rapidly. An assumption is that the target-side of bilingual data plays a more important role in NMT training, thus the relative low-quality target side of FT data hinders the NMT performance. Instead, for using FT data, we firstly train a NMT model with FT data, and then fine-tune the model with the authentic data in all our experiments.} is much lower than BT, which is consistent with the observations in ~\cite{bogoychev2019domain}.

\paragraph{Results of AR Models} 
With respect to BLEU scores, we observe that the *-ADD models work better than the *-REP models. This indicates that both synthetic and repaired data are useful for NMT training. All repaired models (BTR and FTR) achieve better results over corresponding Non-repaired models. The BASE + BTR-ADD + FTR-ADD model achieves the best BLEU score over all test sets. 

On the WMT14 EN-DE translation task, the BASE + BTR-ADD + FTR-ADD model outperforms the baseline by 2.07 BLEU points and outperforms the BASE + BT + FT model by 0.75 BLEU points. While on the IWSLT DE-EN task, The BASE + BTR-ADD + FTR-ADD model outperforms the baseline and the BASE + BT + FT model by 3.85 and 1.19 BLEU points, respectively. An average 2.96 BLEU points improvement over baseline and 0.97 BLEU point improvement over strong synthetic model are achieved by our best AR models.

\subsection{AR Quality Analysis}
\label{sec:arqa}
\begin{table}[!t]
\begin{tabular}{c||c|c|c|c}
\hline
\multirow{2}{*}{Model} & \multicolumn{2}{c|}{BLEU} & \multirow{2}{*}{CR} & \multirow{2}{*}{BR} \\ \cline{2-3}
                       & $S_s$         & $S_{sr}$                                    &                              \\ \hline
EN2EN                  & 47.02       & \textbf{58.47}      & 79.40\%                      & 72.17\%                      \\ \hline
DE2DE                  & 34.71       & \textbf{50.12}     & 76.80\%                      & 69.90\%                      \\ \hline

\end{tabular}
\caption{Quality analysis on the AR models. \textbf{CR} means the change rate. \textbf{BR} indicates the better Rate. Compared with $S_s$, $S_{sr}$ achieve over 11 BLEU points improvement on both EN$\rightarrow$DE and DE$\rightarrow$EN tasks.}
\label{table:analysis}
\end{table}



We further analyze the quality of the proposed AR models themselves from three aspects: BLEU score, Change Rate, Better Rate in Table~\ref{table:analysis}. 
We follow the definition of dev data described in Section~\ref{sec:artd}.
We use our AR model to transform from $S_s$ to $S_{sr}$ of the dev data, ($S_s, S_a$), then use $S_a$ as the reference to evaluate the quality of $S_a$.
\paragraph{BLEU Score} We apply the corpus-level BLEU score as the evaluation criterion to measure whether $S_{sr}$ has higher quality than $S_s$. From Table~\ref{sec:arqa}, we can find that $S_{sr}$ achieve over +11 BLEU points improvements over $S_S$ of both EN2EN and DE2DE AR models. The BLEU score improvements indicate that our AR models can improve the quality of synthetic data.
\paragraph{Change Rate} We count the differences between the $S_s$ and $S_{sr}$ and call it the change rate. The CR could be calculated as: $$CR=\frac{\#Changed}{\#Sentences}$$, 
where $\#Changed$ is the number of sentence that the $S_{sr}$ is different from the input sentence $S_s$. We found that over 76\% of the input sentences have been changed by the AR models, which indicates that our AR models indeed learn transformation information for synthetic data.

\paragraph{Better Rate} "Better" here means $S_{sr}$ achieves a higher sentence-level BLEU score compared with $S_s$. We can find that over 69\% of $S_{sr}$ are better than $S_s$, which proves again the AR models indeed improve the quality of synthetic data. 

\section{Conclusion}
\label{sec:conclusion}
In this paper, we have presented an AR framework by using seq2seq-based AR model to directly \textit{repair} the synthetic parallel data. The proposed method can be applied to various types of synthetic corpus with less time cost. 
On both WMT14 EN$\rightarrow$DE and IWSLT14 DE$\rightarrow$EN translation tasks, experimental results and further in-depth analysis demonstrate that the proposed AR method is able to 1) improve the quality of synthetic parallel data; 2) significantly improve the quality of NMT models by repaired data.


\bibliography{acl2020}

\begin{thebibliography}{25}
\expandafter\ifx\csname natexlab\endcsname\relax\def\natexlab#1{#1}\fi

\bibitem[{Bahdanau et~al.(2014)Bahdanau, Cho, and Bengio}]{bahdanau2014neural}
Dzmitry Bahdanau, Kyunghyun Cho, and Yoshua Bengio. 2014.
\newblock Neural machine translation by jointly learning to align and
  translate.
\newblock \emph{arXiv preprint arXiv:1409.0473}.

\bibitem[{Bogoychev and Sennrich(2019)}]{bogoychev2019domain}
Nikolay Bogoychev and Rico Sennrich. 2019.
\newblock Domain, translationese and noise in synthetic data for neural machine
  translation.
\newblock \emph{arXiv preprint arXiv:1911.03362}.

\bibitem[{Caswell et~al.(2019)Caswell, Chelba, and
  Grangier}]{caswell2019tagged}
Isaac Caswell, Ciprian Chelba, and David Grangier. 2019.
\newblock Tagged back-translation.
\newblock \emph{arXiv preprint arXiv:1906.06442}.

\bibitem[{Cho et~al.(2014)Cho, van Merrienboer, Gulcehre, Bahdanau, Bougares,
  Schwenk, and Bengio}]{Cho2014Learning}
Kyunghyun Cho, Bart van Merrienboer, Caglar Gulcehre, Dzmitry Bahdanau, Fethi
  Bougares, Holger Schwenk, and Yoshua Bengio. 2014.
\newblock Learning phrase representations using rnn encoder--decoder for
  statistical machine translation.
\newblock In \emph{EMNLP}.

\bibitem[{Edunov et~al.(2018)Edunov, Ott, Auli, and
  Grangier}]{edunov2018understanding}
Sergey Edunov, Myle Ott, Michael Auli, and David Grangier. 2018.
\newblock Understanding back-translation at scale.
\newblock \emph{arXiv preprint arXiv:1808.09381}.

\bibitem[{Gao et~al.(2019)Gao, Zhu, Wu, Xia, Qin, Cheng, Zhou, and
  Liu}]{gao2019soft}
Fei Gao, Jinhua Zhu, Lijun Wu, Yingce Xia, Tao Qin, Xueqi Cheng, Wengang Zhou,
  and Tie-Yan Liu. 2019.
\newblock Soft contextual data augmentation for neural machine translation.
\newblock In \emph{Proceedings of the 57th Annual Meeting of the Association
  for Computational Linguistics}, pages 5539--5544.

\bibitem[{Gehring et~al.(2017)Gehring, Auli, Grangier, Yarats, and
  Dauphin}]{gehring2017convolutional}
Jonas Gehring, Michael Auli, David Grangier, Denis Yarats, and Yann~N Dauphin.
  2017.
\newblock Convolutional sequence to sequence learning.
\newblock In \emph{Proceedings of the 34th International Conference on Machine
  Learning-Volume 70}, pages 1243--1252. JMLR. org.

\bibitem[{Gulcehre et~al.(2015)Gulcehre, Firat, Xu, Cho, Barrault, Lin,
  Bougares, Schwenk, and Bengio}]{gulcehre2015using}
Caglar Gulcehre, Orhan Firat, Kelvin Xu, Kyunghyun Cho, Loic Barrault, Huei-Chi
  Lin, Fethi Bougares, Holger Schwenk, and Yoshua Bengio. 2015.
\newblock On using monolingual corpora in neural machine translation.
\newblock \emph{arXiv preprint arXiv:1503.03535}.

\bibitem[{Gwinnup et~al.(2017)Gwinnup, Anderson, Erdmann, Young, Kazi, Salesky,
  Thompson, and Taylor}]{gwinnup2017afrl}
Jeremy Gwinnup, Timothy Anderson, Grant Erdmann, Katherine Young, Michaeel
  Kazi, Elizabeth Salesky, Brian Thompson, and Jonathan Taylor. 2017.
\newblock The afrl-mitll wmt17 systems: Old, new, borrowed, bleu.
\newblock In \emph{Proceedings of the Second Conference on Machine
  Translation}, pages 303--309.

\bibitem[{He et~al.(2016)He, Xia, Qin, Wang, Yu, Liu, and Ma}]{he2016dual}
Di~He, Yingce Xia, Tao Qin, Liwei Wang, Nenghai Yu, Tie-Yan Liu, and Wei-Ying
  Ma. 2016.
\newblock Dual learning for machine translation.
\newblock In \emph{Advances in Neural Information Processing Systems}, pages
  820--828.

\bibitem[{Hoang et~al.(2018)Hoang, Koehn, Haffari, and
  Cohn}]{hoang2018iterative}
Vu~Cong~Duy Hoang, Philipp Koehn, Gholamreza Haffari, and Trevor Cohn. 2018.
\newblock Iterative back-translation for neural machine translation.
\newblock In \emph{Proceedings of the 2nd Workshop on Neural Machine
  Translation and Generation}, pages 18--24.

\bibitem[{Kingma and Ba(2014)}]{kingma2014adam}
Diederik~P Kingma and Jimmy Ba. 2014.
\newblock Adam: A method for stochastic optimization.
\newblock \emph{arXiv preprint arXiv:1412.6980}.

\bibitem[{Lample et~al.(2018)Lample, Ott, Conneau, Denoyer, and
  Ranzato}]{lample2018phrase}
Guillaume Lample, Myle Ott, Alexis Conneau, Ludovic Denoyer, and Marc'Aurelio
  Ranzato. 2018.
\newblock Phrase-based \& neural unsupervised machine translation.
\newblock \emph{arXiv preprint arXiv:1804.07755}.

\bibitem[{Papineni et~al.(2002)Papineni, Roukos, Ward, and
  Zhu}]{papineni2002bleu}
Kishore Papineni, Salim Roukos, Todd Ward, and Wei-Jing Zhu. 2002.
\newblock Bleu: a method for automatic evaluation of machine translation.
\newblock In \emph{Proceedings of the 40th annual meeting on association for
  computational linguistics}, pages 311--318. Association for Computational
  Linguistics.

\bibitem[{Poncelas et~al.(2018)Poncelas, Shterionov, Way, Wenniger, and
  Passban}]{poncelas2018investigating}
Alberto Poncelas, Dimitar Shterionov, Andy Way, Gideon Maillette de~Buy
  Wenniger, and Peyman Passban. 2018.
\newblock Investigating backtranslation in neural machine translation.
\newblock \emph{arXiv preprint arXiv:1804.06189}.

\bibitem[{Sennrich et~al.(2017)Sennrich, Birch, Currey, Germann, Haddow,
  Heafield, Barone, and Williams}]{sennrich2017university}
Rico Sennrich, Alexandra Birch, Anna Currey, Ulrich Germann, Barry Haddow,
  Kenneth Heafield, Antonio Valerio~Miceli Barone, and Philip Williams. 2017.
\newblock The university of edinburgh's neural mt systems for wmt17.
\newblock \emph{arXiv preprint arXiv:1708.00726}.

\bibitem[{Sennrich et~al.(2015)Sennrich, Haddow, and
  Birch}]{sennrich2015neural}
Rico Sennrich, Barry Haddow, and Alexandra Birch. 2015.
\newblock Neural machine translation of rare words with subword units.
\newblock \emph{arXiv preprint arXiv:1508.07909}.

\bibitem[{Sennrich and Zhang(2019)}]{sennrich2019revisiting}
Rico Sennrich and Biao Zhang. 2019.
\newblock Revisiting low-resource neural machine translation: A case study.
\newblock In \emph{Proceedings of the 57th Annual Meeting of the Association
  for Computational Linguistics}.

\bibitem[{Sutskever et~al.(2014)Sutskever, Vinyals, and
  Le}]{sutskever2014sequence}
I~Sutskever, O~Vinyals, and QV~Le. 2014.
\newblock Sequence to sequence learning with neural networks.
\newblock \emph{Advances in NIPS}.

\bibitem[{Vaswani et~al.(2018)Vaswani, Bengio, Brevdo, Chollet, Gomez, Gouws,
  Jones, Kaiser, Kalchbrenner, Parmar et~al.}]{vaswani2018tensor2tensor}
Ashish Vaswani, Samy Bengio, Eugene Brevdo, Francois Chollet, Aidan~N Gomez,
  Stephan Gouws, Llion Jones, {\L}ukasz Kaiser, Nal Kalchbrenner, Niki Parmar,
  et~al. 2018.
\newblock Tensor2tensor for neural machine translation.
\newblock \emph{arXiv preprint arXiv:1803.07416}.

\bibitem[{Vaswani et~al.(2017)Vaswani, Shazeer, Parmar, Uszkoreit, Jones,
  Gomez, Kaiser, and Polosukhin}]{vaswani2017attention}
Ashish Vaswani, Noam Shazeer, Niki Parmar, Jakob Uszkoreit, Llion Jones,
  Aidan~N Gomez, {\L}ukasz Kaiser, and Illia Polosukhin. 2017.
\newblock Attention is all you need.
\newblock In \emph{Advances in neural information processing systems}, pages
  5998--6008.

\bibitem[{Wang et~al.(2019)Wang, Liu, Wang, Luan, and Sun}]{wang2019improving}
Shuo Wang, Yang Liu, Chao Wang, Huanbo Luan, and Maosong Sun. 2019.
\newblock Improving back-translation with uncertainty-based confidence
  estimation.
\newblock \emph{arXiv preprint arXiv:1909.00157}.

\bibitem[{Wu et~al.(2019)Wu, Wang, Xia, Tao, Lai, and Liu}]{wu2019exploiting}
Lijun Wu, Yiren Wang, Yingce Xia, QIN Tao, Jianhuang Lai, and Tie-Yan Liu.
  2019.
\newblock Exploiting monolingual data at scale for neural machine translation.
\newblock In \emph{Proceedings of the 2019 Conference on Empirical Methods in
  Natural Language Processing and the 9th International Joint Conference on
  Natural Language Processing (EMNLP-IJCNLP)}, pages 4198--4207.

\bibitem[{Wu et~al.(2016)Wu, Schuster, Chen, Le, Norouzi, Macherey, Krikun,
  Cao, Gao, Macherey et~al.}]{wu2016google}
Yonghui Wu, Mike Schuster, Zhifeng Chen, Quoc~V Le, Mohammad Norouzi, Wolfgang
  Macherey, Maxim Krikun, Yuan Cao, Qin Gao, Klaus Macherey, et~al. 2016.
\newblock Google's neural machine translation system: Bridging the gap between
  human and machine translation.
\newblock \emph{arXiv preprint arXiv:1609.08144}.

\bibitem[{Xia et~al.(2019)Xia, Tan, Tian, Gao, He, Chen, Fan, Gong, Leng, Luo,
  Wang, Wu, Zhu, Qin, and Liu}]{xia2019microsoft}
Yingce Xia, Xu~Tan, Fei Tian, Fei Gao, Di~He, Weicong Chen, Yang Fan, Linyuan
  Gong, Yichong Leng, Renqian Luo, Yiren Wang, Lijun Wu, Jinhua Zhu, Tao Qin,
  and Tie-Yan Liu. 2019.
\newblock {M}icrosoft research asia{'}s systems for {WMT}19.
\newblock In \emph{the Fourth Conference on Machine Translation)}.

\end{thebibliography}
\bibliographystyle{acl_natbib}

\end{document}